\documentclass[letterpaper]{article} 
\usepackage{aaai25}  
\usepackage{times}  
\usepackage{helvet}  
\usepackage{courier}  
\usepackage[hyphens]{url}  
\usepackage{graphicx} 
\usepackage{natbib}  
\usepackage{caption} 
\usepackage{algorithm}
\usepackage{algorithmic}

\usepackage{subcaption}
\usepackage{cuted}
\usepackage{amssymb}
\usepackage{amsmath}
\usepackage{mathrsfs}
\usepackage{booktabs}
\usepackage{multirow}
\usepackage{adjustbox}
\usepackage{tabularx}
\usepackage[table]{xcolor}
\usepackage{makecell}
\DeclareMathOperator*{\argmin}{arg\,min}
\usepackage{array}

\usepackage{newfloat}
\usepackage{listings}
\DeclareCaptionStyle{ruled}{labelfont=normalfont,labelsep=colon,strut=off} 
\floatstyle{ruled}
\newfloat{listing}{tb}{lst}{}
\floatname{listing}{Listing}
\title{DriveEditor: A Unified 3D Information-Guided Framework \\ for Controllable Object Editing in Driving Scenes}
\author {
    Yiyuan Liang\textsuperscript{\rm 1, \rm 2}\equalcontrib,
    Zhiying Yan\textsuperscript{\rm 1, \rm 2}\equalcontrib,
    Liqun Chen\textsuperscript{\rm 1, \rm 2}\equalcontrib,
    Jiahuan Zhou\textsuperscript{\rm 3}, \\
    Luxin Yan\textsuperscript{\rm 1, \rm 2},
    Sheng Zhong\textsuperscript{\rm 1, \rm 2},
    Xu Zou\textsuperscript{\rm 1, \rm 2}\thanks{Corresponding author.}
}
\affiliations {
    \textsuperscript{\rm 1}Huazhong University of Science and Technology\\
    \textsuperscript{\rm 2}National Key Laboratory of Multispectral Information Intelligent Processing Technology\\
    \textsuperscript{\rm 3}Wangxuan Institute of Computer Technology, Peking University\\
    \{yvanliang, yanzhiying, chenliqun, yanluxin, zhongsheng, zoux\}@hust.edu.cn, jiahuanzhou@pku.edu.cn
}

\begin{document}

\maketitle

\begin{figure*}[h!]
\centering
\includegraphics[width=0.958\textwidth]{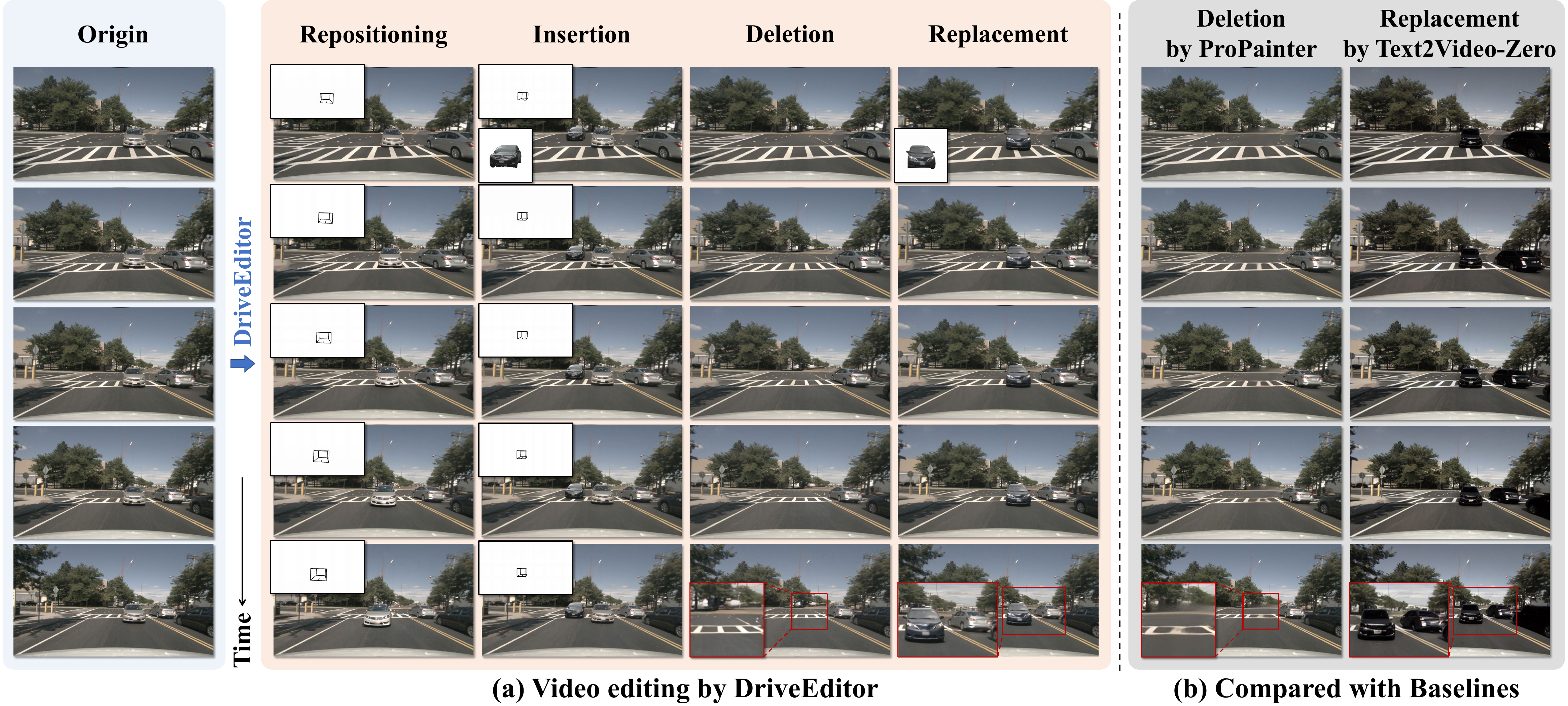}
\captionof{figure}{\textbf{Visualizations of the editing capability of DriveEditor and baselines. (a)} DriveEditor enables user-friendly \textit{repositioning}, \textit{insertion}, \textit{replacement}, and \textit{deletion} within a unified framework. It precisely controls an object's position and orientation based on the 3D bounding box (top left; required for repositioning and insertion tasks that alter object position), and maintains high-fidelity appearance attributes of the object from a single reference image (bottom left, required for insertion and replacement tasks that alter object appearance). \textbf{(b)} The \textit{deletion} and \textit{replacement} results compared with baselines. ProPainter's deletion results suffer from artifacts. Text2Video-Zero employs the text prompt ``replace the champagne-colored car with dark gray van'' to guide the replacement process. Yet, it produces unrealistic visual results and alters the appearance of other vehicles.}
\label{fig1}
\end{figure*}

\begin{abstract}
Vision-centric autonomous driving systems require diverse data for robust training and evaluation, which can be augmented by manipulating object positions and appearances within existing scene captures. While recent advancements in diffusion models have shown promise in video editing, their application to object manipulation in driving scenarios remains challenging due to imprecise positional control and difficulties in preserving high-fidelity object appearances. To address these challenges in position and appearance control, we introduce DriveEditor, a diffusion-based framework for object editing in driving videos. DriveEditor offers a unified framework for comprehensive object editing operations, including repositioning, replacement, deletion, and insertion. These diverse manipulations are all achieved through a shared set of varying inputs, processed by identical position control and appearance maintenance modules. The position control module projects the given 3D bounding box while preserving depth information and hierarchically injects it into the diffusion process, enabling precise control over object position and orientation. The appearance maintenance module preserves consistent attributes with a single reference image by employing a three-tiered approach: low-level detail preservation, high-level semantic maintenance, and the integration of 3D priors from a novel view synthesis model. Extensive qualitative and quantitative evaluations on the nuScenes dataset demonstrate DriveEditor's exceptional fidelity and controllability in generating diverse driving scene edits, as well as its remarkable ability to facilitate downstream tasks.
\end{abstract}

\begin{links}
\link{Code}{https://github.com/yvanliang/DriveEditor}
\end{links}

\section{Introduction}
In autonomous driving, perception tasks~\cite{bevfusion, StreamPETR, FocalFormer3D, cvt} necessitate extensive data to construct robust models. To facilitate this, large-scale, open-source datasets~\cite{nuscenes, waymo, once} for autonomous driving have been introduced, which contain thousands of driving scenes. While these datasets offer a rich repository of road-collected driving data, they exhibit a long-tailed distribution if not further processed. This imbalance leads to an overrepresentation of common driving scenes and an underrepresentation of rare yet critical events, such as unexpected obstacles or lane changes, posing challenges for training and evaluating perception tasks under these scenarios.

To mitigate this diversity challenge in driving data, recent approaches~\cite{drivedreamer, delphi, magicdrive} have leveraged the capabilities of Latent Diffusion Models~(LDMs)~\cite{ldm} to generate a variety of driving scenes. By leveraging Bird's-Eye-View (BEV) layouts to constrain scene structure, including lane lines and object positions, these methods generate diverse scenes aligned with semantic driving scenarios. However, they offer limited control over objects, operating at a semantic level without fine-grained constraints on detailed appearance. Despite using a subject bank for object control, SubjectDrive~\cite{subjectdrive} still offers limited control over fine-grained object details. Beyond video generation, LDMs have also demonstrated remarkable capabilities in video editing. Leveraging natural language descriptions, some artworks \cite{videdit, realcraft, store, Edit-A-Video} achieve content alteration and style transfer within videos, producing remarkable outcomes. Nonetheless, these methods struggle to dictate precise visual appearances due to inherent linguistic ambiguities, and they also find it difficult to change object positions, which hinders object editing in driving contexts.

This paper introduces DriveEditor, a unified diffusion-based framework designed for the precise manipulation of objects within driving scenario videos. While seamlessly preserving the original background, DriveEditor enables a comprehensive set of object editing operations, including repositioning, replacement, deletion, and insertion, as illustrated in Figure~\ref{fig1}. Notably, these diverse manipulations are all achieved via a shared set of varying inputs processed by two core modules: the position control module and the appearance maintenance module. The position control module projects each face of the 3D bounding box individually onto the image plane, preserving depth information. This projected data is then hierarchically injected into the diffusion process, enabling control over object position and orientation. To guarantee faithful visual appearance consistency with the single reference image, we introduce three distinct levels of appearance control: low-level details preservation, high-level semantics maintenance, and incorporation of 3D priors derived from the novel view synthesis model.

DriveEditor comprehensively covers a wide range of object editing operations, significantly enriching the diversity of autonomous driving data. On one hand, it operates in a user-friendly manner, requiring only a single reference image and 3D bounding boxes per frame, which can be easily obtained through our interactive tool. On the other hand, it offers high controllability, enabling precise adjustments to object appearance, position, and orientation based on user instructions, or through automatic means.

Our main contributions can be summarized as follows:
\begin{itemize}
    \item We propose DriveEditor, a diffusion-based framework for object editing in videos of driving scenarios.
    \item DriveEditor accomplishes four distinct editing tasks, leveraging shared inputs and a common network to enhance each task.
    \item We demonstrate DriveEditor's exceptional fidelity, controllability, and efficacy in facilitating downstream tasks through comprehensive experiments.
\end{itemize}

\section{Related Works}
\begin{figure*}[t]
  \centering
  \includegraphics[width=0.996\textwidth]{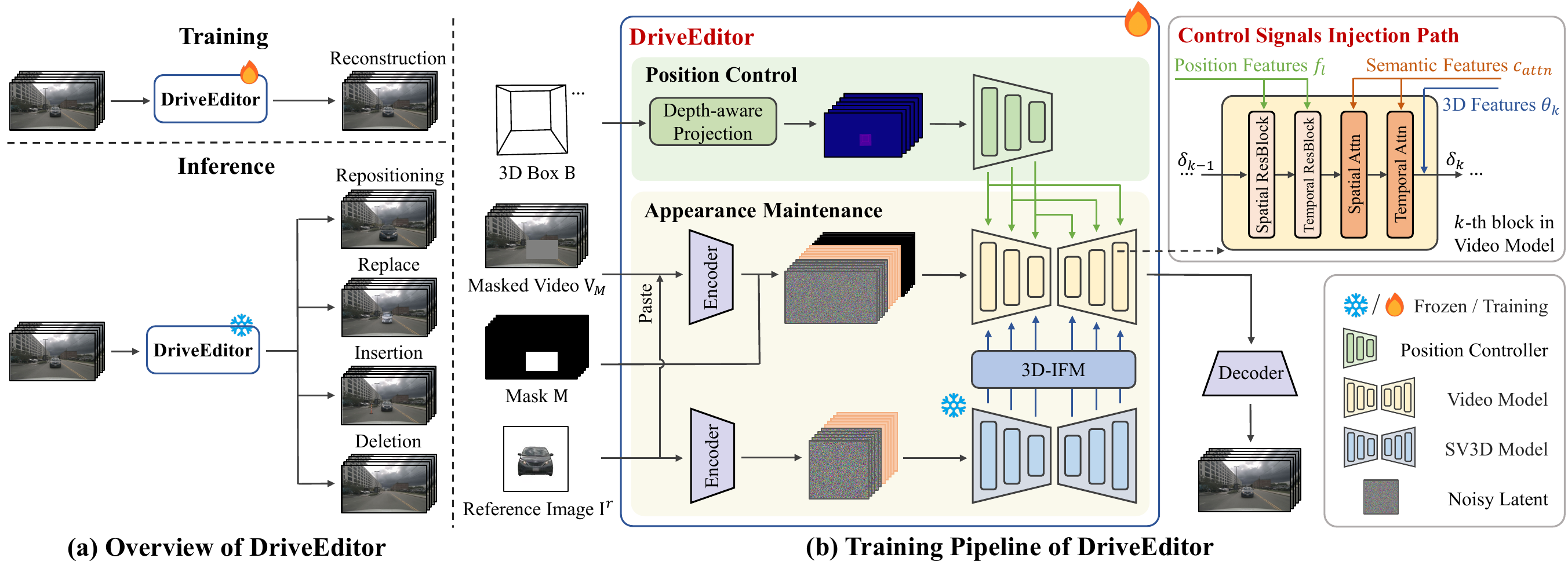}
  \caption{\textbf{(a)} High-level overview of DriveEditor. \textbf{(b)} Diagram of the training pipeline of DriveEditor. Three levels of appearance control are established based on the single reference image $\textbf{I}^r$: low-level details preservation through a cut-and-paste approach, high-level semantics maintenance through cross-attention (omitted in the pipeline for brevity), and incorporation of 3D priors derived from the frozen SV3D U-Net. For position control, we perform a projection that preserves depth information, followed by the Pose Controller to extract multi-scale features. Control signals are injected through three distinct paths in block of the video model: position features into ResBlocks, semantic features via cross-attention, and 3D features added to block outputs.}
  \label{fig2}
\end{figure*}

\noindent\textbf{Driving Scene Manipulation.} The increasing demand for driving scenario data, coupled with the high cost of detailed manual annotation, highlights the need for efficient methods of acquiring driving scenes. Leveraging the powerful generative capabilities of LDMs, several approaches~\cite{bevcontrol, drivingdiffusion, panacea} generate driving scenes with rich diversity. However, due to the complexity of driving scenarios and the multitude of targets, generated scenes often suffer from quality issues and lack fine-grained control over specific attributes or color variations of individual objects. Other NeRF-based~\cite{nerf} artworks~\cite{chatsim, unisim} enable driving scene reconstruction, facilitating scene simulation and editing. These methods can produce high-fidelity scenes and support viewpoint changes. However, due to limited generative capabilities, they lack the flexibility for object-level editing, such as the removing static objects.

\noindent\textbf{Diffusion Models for Video Editing.} Video editing involves manipulating foreground, background, or style within videos, guided by a target prompt. Tune-A-Video~\cite{tune-a-video} introduces one-shot tuning for video editing by adapting spatial self-attention layers in text-to-image diffusion models to their spatial-temporal counterparts. However, this approach incurs substantial fine-tuning costs. To address this, other methods~\cite{fatezero, infusion, zeroshotvideoeditingusing} enable zero-shot video editing by obtaining hidden features of the original video through inversion and preserving the information via attention map injection. To achieve finer-grained control, recent approaches leverage control modules like ControlNet~\cite{controlnet} to guide editing using multimodal conditions such as depth~\cite{eve}, sketches~\cite{rerender_a_video}, images~\cite{moonshot} and keypoints~\cite{realcraft}, enabling diverse editing instructions.

\section{Data Construction}
DriveEditor is trained via a \textit{reconstruction} task. In this task, we occlude a specific object in a video sequence. DriveEditor is trained to reconstruct the occluded object given a single image and ground-truth bounding boxes of that object. As no existing dataset directly caters to our editing requirements, we construct a dataset based on the widely-used nuScenes dataset~\cite{nuscenes}. This large-scale dataset for autonomous driving comprises 1,000 driving scenes, each scene consists of 20 seconds of videos captured from six camera views at 10 Hz.

Specifically, for each object that remains unobstructed within a 20-meter radius of the camera across $N$ consecutive frames, those frames are concatenated to form an original video $\textbf{V} \in \mathbb{R}^{N \times 3\times H\times W}$. To obtain object images for editing, we employ SAM~\cite{sam} to extract the object from each frame of $\textbf{V}$. These images serve as both training data and an object bank for object replacement. To enhance the model's generalization ability, we randomly select the $r$-th image as the reference image, denoted as $\textbf{I} \in \mathbb{R}^{3\times h\times w}$. We apply masks $\textbf{M}$ to the object region in the video, creating a masked video $\textbf{V}_m$. $\textbf{B}=\left\{ \left ( x_i, y_i, z_i \right)^j \right\} \in \mathbb{R} ^ {N \times 8 \times 3}$ represents the 3D bounding box of the object in the camera coordinate system. The camera's elevation and azimuth angles relative to the object, denoted by $\textbf{e} \in \mathbb{R}^N$ and $\textbf{a} \in \mathbb{R}^N$ respectively, can be calculated based on the center point of $\textbf{B}$ and the camera's position in the world coordinate system. To enhance the model's ability to remove objects, we additionally apply random masks to object-free regions in the videos, generating inpainting training data.

This process yields a training dataset of 10,110 video clips, including a dedicated inpainting training subset of 2,000 clips, and a validation set of 800 clips.

\section{Method}
\subsection{Brief Introduction of Video Diffusion Models}
\noindent\textbf{Stable Video Diffusion~(SVD)}~\cite{svd} is a latent diffusion model specialized in high-quality image-to-video generation. Given a reference image $\mathbf{I} \in \mathbb{R}^{3\times H\times W}$, SVD can generate a video sequence $\mathbf{V} \in \mathbb{R}^{N \times 3 \times H \times W}$ consisting of $N$ frames, initiated from $\mathbf{I}$. Following the EDM-preconditioning framework~\cite{edm}, SVD employs a learnable denoiser $D_\mathbf{\theta}$ to parameterize the U-Net~\cite{unet} network, iteratively estimating $\mathbf{z_0}$, the latent representation of $\mathbf{V_0}$, from $\mathbf{z_M} \sim \mathcal{N}(\mathbf{0}, {\sigma_{max}}^2)$ through denoising score matching:\begin{equation}\fontsize{9}{0}\mathbb{E} _{\mathbf{z}_0 \sim p_{data} , \left( \sigma, \mathbf{n} \right ) \sim p \left( \sigma, \mathbf{n} \right )}\left [ \lambda_\sigma \left\| D _\mathbf{\theta} \left ( \mathbf{z}_0 + \mathbf{n}; \sigma, \mathbf{y}, \mathbf{c} \right ) - \mathbf{z}_0 \right\| _{2}^{2} \right ],\end{equation} where $p \left( \sigma, \mathbf{n} \right )=p \left( \sigma \right ) \mathcal{N} \left( \mathbf{n};\mathbf{0}, \sigma^2 \right )$, $\sigma$ denotes the noise level. $\lambda _ \sigma : \mathbb{R}_{+}\to \mathbb{R}_{+}$ is a weighting function. Vector conditionings $\mathbf{y}$ (e.g., fps, motion rate), along with $\sigma$, are embedded and injected into the ResBlocks in the U-Net for guidance. The control signal $\mathbf{c}$ comprises tokens generated by a CLIP~\cite{clip} image encoder from $x^0$, and latent representation produced by a VAE encoder. These are integrated into the diffusion model via cross-attention and channel-wise concatenation with frame latents, respectively.

\noindent\textbf{Stable Video 3D~(SV3D)}~\cite{sv3d} represents a state-of-the-art latent video diffusion model capable of generating multi-view orbital videos around a 3D object. SV3D takes a single object image $\mathbf{I} \in \mathbb{R}^{3\times H\times W}$ as the initial viewpoint and generates an orbital video $\mathbf{V} \in \mathbb{R}^{N \times 3 \times H \times W}$ of the object by controlling the camera pose trajectory $\boldsymbol{\pi}=\left\{\left ( e^{i}, a^{i} \right ) \right\}_{i=1}^{N} \in  \mathbb{R} ^{N \times 2}$ through its vector conditionings $\mathbf{y}$. Here, $e$ represents the elevation angle, and $a$ represents the azimuth angle.

\begin{figure}[t]
\centering
\includegraphics[width=\columnwidth]{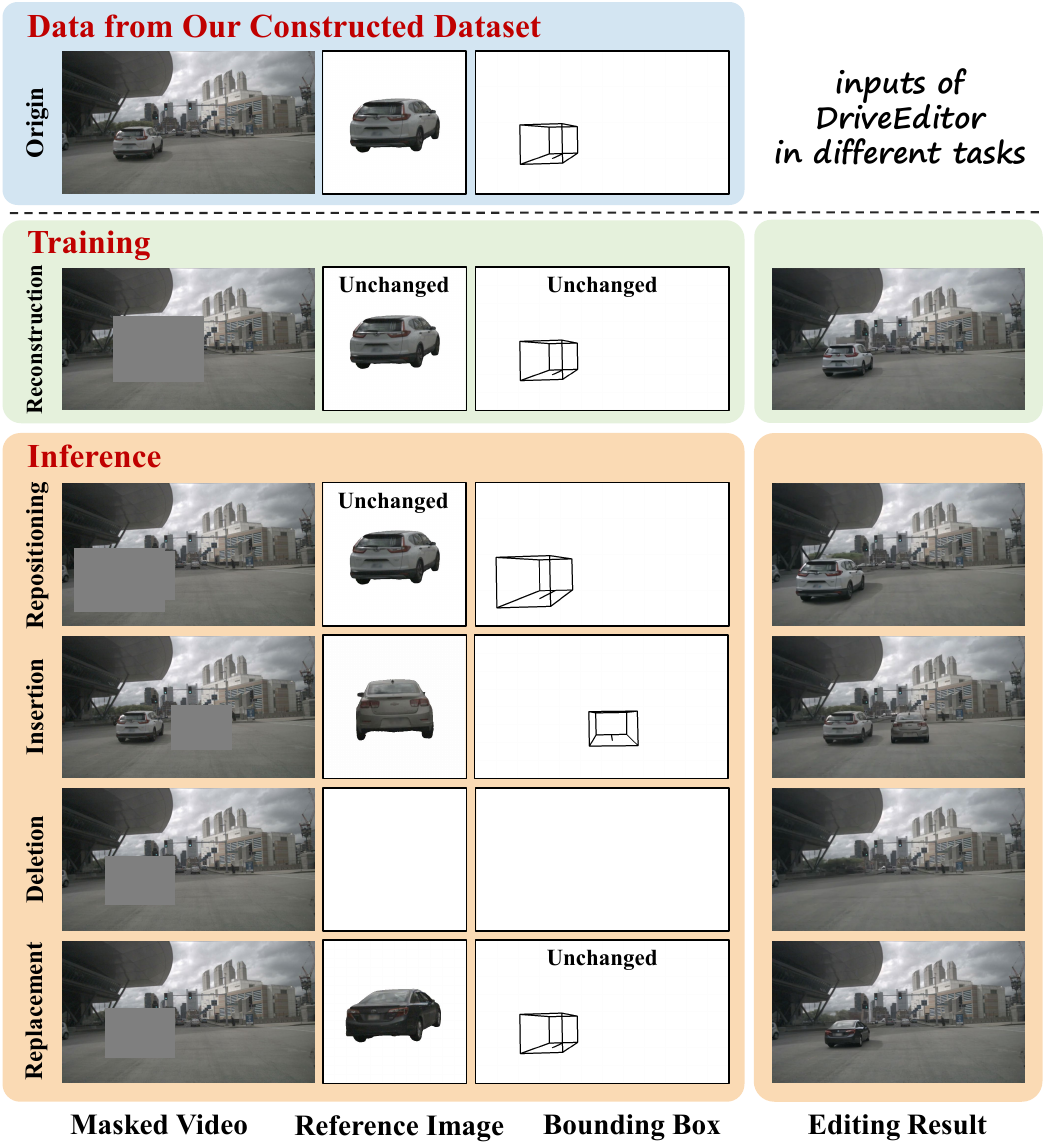}
\caption{DriveEditor is trained to reconstruct occluded objects using inputs from our dataset. At inference time, it performs various editing tasks based on specific input prompts.}
\label{fig3}
\end{figure}

\subsection{Unlocking Unified Editing}
Here we explain the underlying principles that enable seamless repositioning, insertion, replacement, and deletion within DriveEditor framework through a unified pipeline.

During training, masks $\textbf{M}$ are randomly sized and positioned, preventing the model from relying on any visual cues. DriveEditor reconstructs the masked area to recover the object, leveraging appearance cues from $\textbf{I}$ and location information from $\textbf{B}$. It is also trained on inpainting data to restore masked backgrounds, facilitating object deletion.

During inference, DriveEditor leverages varying inputs to facilitate a variety of editing tasks as illustrated in Figure~\ref{fig3}. 

\noindent\textbf{Repositioning} The object is segmented from the original video, obtaining $\textbf{I}^{\prime}$ as a reference to preserve its appearance. The desired position $\textbf{B}^{\prime}$, is controlled by the user. A mask is applied to the original video, encompassing the union of the projected areas defined by the original and desired positions, $\textbf{B}$ and $\textbf{B}^{\prime}$, respectively, to derive $\textbf{V}_m$.

\noindent\textbf{Insertion.} The desired object image and its target bounding box denoted as $\textbf{I}^{\prime}$ and $\textbf{B}^{\prime}$, respectively, are provided as user inputs. $\textbf{V}_m$ is generated within the projected areas of $\textbf{B}^{\prime}$.

\noindent\textbf{Deletion.} As no target object exists, $\textbf{I}^{\prime}$ and $\textbf{B}^{\prime}$ are simply omitted. $\textbf{V}_m$ is generated within the projected areas of $\textbf{B}$.

\noindent\textbf{Replacement.} The user provides a reference image $\textbf{I}^{\prime}$ from the same viewpoint as in the video. If the viewpoints differ, SV3D model can be used for alignment. $\textbf{V}_m$ is generated using the same strategy as in the deletion process.

Finally, the derived reference image, 3D bounding box, and masked video are input to DriveEditor for object editing.

\subsection{Appearance Control}
To control the appearance of objects using the reference image, we employ a three-level approach that facilitates details preservation, semantics maintenance, and 3D priors incorporation.

\noindent\textbf{Low-Level Details Preservation.} To preserve fine-grained details from the reference image, we adopt a straightforward yet efficient cut-and-paste approach. During training, the reference image $\textbf{I}$ is directly pasted onto the $r$-th frame of $\textbf{V}_m$, denoted as $\textbf{V}_m^r$, as they share the same viewpoint. During inference, the provided reference image $\textbf{I}^{\prime}$ is pasted onto the frame in $\textbf{V}_m$ with the closest viewpoint. We determine the size and location of the object in the $r$-th frame of the video based on the projection of $\textbf{B}^r$. The object in $\textbf{I}$ is then pasted onto the corresponding region in $\textbf{V}_m^r$ to preserve fine-grained details of the object's appearance. The pasted video is concatenated with the masks $\mathbf{M}$ and derive $c_{concat}$, which is then injected into the diffusion process through channel-wise concatenation with the latent representation.

\noindent\textbf{High-Level Semantics Maintenance.} The pioneering work~\cite{paint_by_example} demonstrated the CLIP image encoder's ability to extract high-level semantic image information. We leverage this capability to preserve visual concepts by injecting CLIP features of the target object and origin background scene, denoted as $c_{attn}$, into the video U-Net via cross-attention. The $c_{attn}$ is obtained as follows: \begin{equation} c_{attn} = \left[ \text{CLIP} \! \left (\textbf{V}_m^r \right ), \text{CLIP} \! \left ( \textbf{I} \right )\right ], \end{equation} where $\text{CLIP} \! \left( \cdot \right)$ is the CLIP image encoder, $\left [ \cdot \right ]$ represents channel-wise concatenation operation. For the deletion task, we employ a trainable null embedding as a replacement for the CLIP image features to guide object removal.

\noindent\textbf{3D Prior Incorporation.} While relying on a single reference image simplifies the usage of DriveEditor, it limits access to multi-view object information. To address this limitation, we leverage the SV3D model, which is capable of generating novel views of objects. Sharing the same architecture and latent space with SVD, we seamlessly integrate its intermediate features to guide the appearance of objects during video editing.

Starting from the reference image as an initial view, we employ a pretrained SV3D model to generate $M$ novel object views. Intermediate features extracted from each block of the SV3D model are then injected into the corresponding block of the video model as strong object priors. To address the SV3D model's limitations in trajectory control, especially for fine-grained view changes and non-full-circle trajectories, we use a carefully crafted fixed azimuth angle $\tilde{\textbf{a}} \in \mathbb{R}^M$ during generation (details in supplementary materials). Given the relatively consistent elevation angle across the video, we set $\tilde{\textbf{e}} \in \mathbb{R}^M$ to its mean value. For the $i$-th video frame, we identify the corresponding $j$-th frame in the generated novel views based on its azimuth angle $\textbf{a}^i$: \begin{equation} j = \argmin_{i \in \left\{0, 1, \ldots, N-1 \right\}} \left| \tilde{\textbf{a}}^j - \textbf{a}^i \right|. \end{equation}

To address discrepancies in scale and position between the object in the video and the reference image, we introduce a 3D Information Fusion Module (3D-IFM) for feature alignment and fusion. Specifically, $\boldsymbol{\delta}_{k}^{i}$, $\boldsymbol{\theta}_{k}^{i}$ indicates the feature representations of the $k$-th block for the $i$-th frame extracted from the U-Net of the video model, SV3D model, respectively. We firstly transform $\boldsymbol{\theta}_{k}^{j}$ to align with the scale and position of the object in the video, as indicated by the projection of $\textbf{B}^i$. Similar to ControlNet~\cite{controlnet}, we then employ zero convolution layers to prevent the pretrained video model from being corrupted by noise during the initial training phase. The 3D-IFM serves as a layer-wise intermediary between the SV3D model and the video model, facilitating hierarchical refinement of the representation. The fusion process can be formulated as: \begin{equation} \boldsymbol{\delta}_{k}^{i} = \boldsymbol{\delta}_{k}^{i} + \textbf{M}^{i} \times \mathcal{Z} \! \left( \mathcal{T}_{\textbf{B}^i} \! \left( \boldsymbol{\theta}_{k}^{j} \right) \right), \end{equation} where $\mathcal{T}_{\textbf{B}^i}$ represents the transformation and resizing operation based on $\textbf{B}^i$, $\mathcal{Z}$ denotes a zero convolution layer with weights and biases initialized to $\mathbf{0}$, and $k \in \left\{0, 1, \ldots, 2L-1 \right\}$ refers to the $2L$ blocks in the U-Net, with the first $L$ blocks in the encoder and the last $L$ in the decoder.

\subsection{Position Control}
To precisely control object position and orientation, we introduce the Depth-aware Position Controller, comprising Depth-aware Projection and Position Controller. 

In the Depth-aware Projection module, each face of the 3D bounding box $\textbf{B}$ is processed individually. For each face, $P$ points are interpolated from its four vertices. These face and edge points are projected onto the 2D image plane using camera intrinsics, with their corresponding depth values ($z$) assigned as pixel intensities. This process generates a six-channel depth-aware pose image $\textbf{I}_P$, preserving depth information without inter-face occlusion.

$\textbf{I}_P$ is fed into a ResNet-like Position Controller to extract multi-scale positional control features. This Position Controller contains $L$ blocks, aligning with the hierarchical structure of the video U-Net's encoder and decoder. These features are injected into the spatial and temporal ResBlocks of the video U-Net using adapters. Let $\mathbf{f}_k$ denote the feature at the $k$-th block of the Object Position Controller, and $\mathbf{g}_k \in \mathbb{R}^{N \times C_k\times H_k\times W_k}$ and $\mathbf{v}_k \in \mathbb{R}^{C_k \times N\times H_k\times W_k}$ represent the  spatial and temporal ResBlocks of the $k$-th block in video U-Net, respectively. The injection of position control features can be represented by: \begin{equation} \begin{split} \mathbf{g}_k &= \mathbf{g}_k + \text{Adapter2d}_k\!\left( \mathbf{f}_{l} \right), \\ \mathbf{v}_k &= \mathbf{v}_k + \text{Adapter3d}_k\!\left( \text{Permute} \! \left(\mathbf{f}_{l} \right)\right), \end{split} \end{equation} where $l=k$ if $l < L$ else $2L-k-1$, $\text{AdapterNd} \! \left( \cdot \right)$ represents the $k$-th adapter composed of SiLU, LayerNorm, and ConvNd operations. $\text{Permute} \! \left( \cdot \right)$ denotes the permutation of frame and channel dimensions. By aligning spatially with video model features and ensuring temporal coherence via temporal injection, it facilitating precise position control.

To prevent object location information leakage through the 3D-IFM, we adopt a two-stage training strategy. In the first stage, by training the video model and Position Controller without 3D information fusion, we restrict the model to learning object positions solely from the input, which effectively facilitates the training of the Position Controller. In the second stage, we reintroduce the 3D-IFM and the frozen SV3D model, allowing for the training of the video model alongside all proposed modules.

\section{Experiments}
\begin{figure*}[!t]
  \centering
  \includegraphics[width=\textwidth]{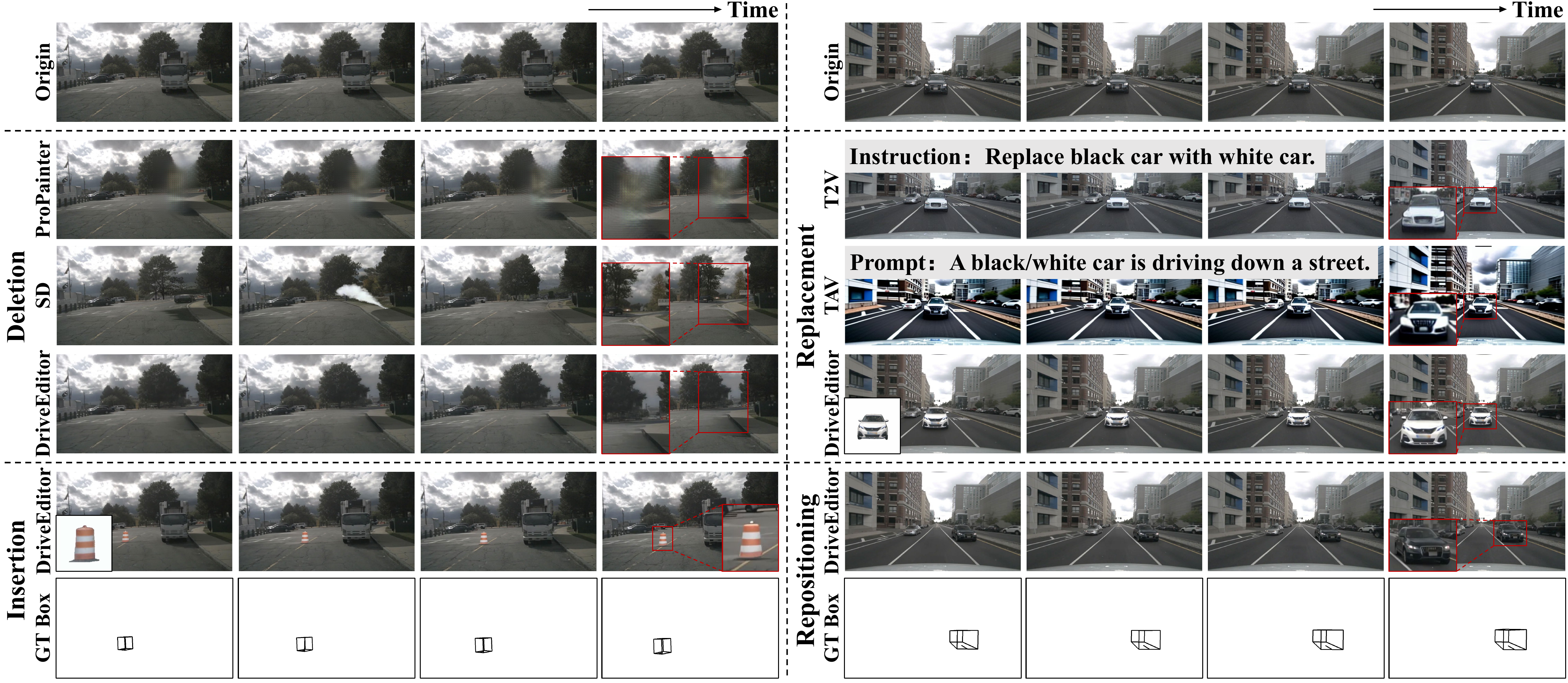}
  \caption{\textbf{Top row:} Original videos. 
  \textbf{Middle left:} Qualitative comparison on the \textit{deletion} task. ProPainter suffers from artifacts, while SD lacks temporal consistency. DriveEditor effectively generates plausible occluded regions.
  \textbf{Middle Right:} Qualitative comparison on the \textit{replacement} task. T2V loses realism, for instance, the roof color remains unchanged. TAV alters the overall style of the video and leads to object deformations. In contrast, DriveEditor maintains high-fidelity object details from the reference image.
  \textbf{Bottom Left:} Visualization of object \textit{insertion} using DriveEditor. It enables precise control over object insertion position while maintaining appearance from the reference image.
  \textbf{Bottom Right:} Visualization of object \textit{repositioning} using DriveEditor. The object is accurately repositioned to align with the GT bounding box while preserving its original appearance.}
  \label{fig4}
\end{figure*}

\subsection{Experimental Setups}
\noindent\textbf{Baselines.} While our work focuses on constructing a unified framework for versatile editing tasks, there is limited related work available, and even fewer methods allow for fair comparison. For \textit{replacement}, we compare our method to Tune-A-Video (T2V)~\cite{tune-a-video} and Text2Video-Zero (T2V)~\cite{Text2Video-Zero}, which excel at video context editing via text prompts. For \textit{deletion}, we compare against ProPainter~\cite{propainter} and Stable Diffusion~\cite{ldm} Inpainting~(SD). To the best of our knowledge, existing video editing methods lack the capability for fine-grained 3D object position manipulation, as they are primarily designed for broader editing tasks. We thus provide quantitative and qualitative results of DriveEditor on \textit{insertion} and \textit{replacement} tasks.

\noindent\textbf{Quality Metrics.} We utilize frame-wise FID~\cite{fid} and FVD~\cite{fvd} metrics to evaluate both image quality and temporal consistency. Additionally, we apply CLIP-I metrics to assess the semantic alignment between the reference image and the object region in the edited video. Given the availability of ground truth~(GT) videos in the reconstruction task, we further employ PSNR and LPIPS~\cite{lpips} to measure pixel-level fidelity, and perceptual differences, respectively.

\noindent\textbf{Position Controllability Metrics.} We assess the alignment between the edited object and the ground truth 3D bounding box using a pretrained StreamPETR~\cite{StreamPETR} model, which is a state-of-the-art multi-view 3D object detector. We use mean Recall (mRecall), mean Average Translation Error (mATE), and mean Average Orientation Error (mAOE) to quantify positional accuracy of editing results. To ensure a fair evaluation of object editing capabilities, our assessment focuses exclusively on the edited objects. 

\noindent\textbf{Model Setup.} DriveEditor generates high-resolution videos of $576\times1024$ pixels with a length of 10 frames. Additional details can be found in the supplementary materials.

\subsection{Main Results}
\noindent\textbf{Editing Quality.} We conduct object editing on the validation set and present both quantitative metrics of editing quality in Table~\ref{tab:1} and qualitative visualizations in Figure~\ref{fig4}. For deletion, ProPainter exhibits severe artifacts, resulting in a lack of realism as indicated by an FID of 36.13. While SD achieves higher image realism with an FID of 30.89, it suffers from inconsistent frame-to-frame results, leading to a higher FVD of 457. In contrast, DriveEditor generates temporally coherent high-quality results, with FID and FVD scores of 30.17 and 228, respectively. For replacement, T2V achieves modifications to object colors but loses detailed realism. TAV alters the overall style and introduces object deformations, resulting in significantly higher FID and FVD scores. In contrast, DriveEditor preserves high-fidelity object details from the reference image, achieving much lower FID and FVD scores of 11.83 and 39, respectively. Visualization results also demonstrate that DriveEditor can produce visually harmonious editing results, while highly preserving the object's appearance attributes.

\begin{table}[t]\small
  \centering
  \setlength{\tabcolsep}{1mm}{
    \begin{tabular}{p{0.21\linewidth}|p{0.22\linewidth}|>{\centering\arraybackslash}p{0.21\linewidth} >{\centering\arraybackslash}p{0.21\linewidth}}
      \toprule
      Tasks & Methods & FID $\downarrow$ & FVD $\downarrow$ \\
      \midrule[0.5pt]
      \multirow{3}[0]{*}{\centering Deletion} & ProPainter & 36.13 & 365\\
      & SD & 30.89 & 457\\
      & DriveEditor & \textbf{30.17} & \textbf{228}\\
      \midrule[0.5pt]
      \multirow{3}[0]{*}{\centering Replacement} & T2V & 14.22 & 151\\
      & TAV & 58.25 & 783\\
      & DriveEditor & \textbf{11.83} & \textbf{39}\\
      \bottomrule
    \end{tabular}
  }
  \caption{Comparison of generation fidelity on replacement and deletion tasks. The best results are in \textbf{bold}. DriveEditor outperforms all baselines in terms of both single-frame generation quality and temporal consistency by a large margin.}
  \label{tab:1}
\end{table}

\begin{table}[t]\small
  \centering
  \setlength{\tabcolsep}{1mm}{
    \begin{tabular}{l@{ }|ccc|c}
      \toprule
      Tasks & \ \ mRecall $\uparrow$ & \ \ mATE $\downarrow$ & \ \ mAOE $\downarrow$ & \ \ CLIP-I $\uparrow$\\
      \midrule[0.5pt]
      Repositioning & \ \ 0.93 & \ \ 0.68 & \ \ 0.044 & \ \ 77.73 \\
      Insertion & \ \ 0.94 & \ \ 0.66 & \ \ 0.043 & \ \ 77.91 \\
      Replacement & \ \ 0.94 & \ \ 0.74 & \ \ 0.044 & \ \ 77.86 \\
      \midrule[0.5pt]
      Oracle & \ \ 0.99 & \ \ 0.42 & \ \ 0.037 & \ \ 78.06 \\
      \bottomrule
    \end{tabular}
  }
  \caption{Evaluation of DriveEditor's positional control via 3D object detection and the semantic alignment between the reference image and the edited object. The performance on the edited results shows only slight degradation compared to the unedited data \textit{oracle}, indicating robust positional control and semantic alignment across all tasks.}
  \label{tab:2}
\end{table}

\begin{table*}[t]\small
  \centering
  \setlength{\tabcolsep}{1mm}{
\begin{tabular}{ccc|cccc|ccc}
    \toprule
\multicolumn{3}{c|}{Settings} & \multicolumn{4}{c|}{Quality} & \multicolumn{3}{c}{Position} \\
    \midrule[0.5pt]
 PC & DP & 3D-IFM& FID $\downarrow$ & FVD $\downarrow$ & PSNR $\uparrow$ & LPIPS $\downarrow$ & mRecall $\uparrow$ & mATE $\downarrow$ & mAOE $\downarrow$ \\
    \midrule[0.5pt]
 - & - & - & 6.56 & 24.95 & 29.83 & 0.052 & 0.91 & 0.74 & 0.197 \\
\checkmark& - & - & 6.59 & 25.29 & 29.82 & 0.052 & 0.92 & 0.71 & 0.148 \\
\checkmark&\checkmark& - & 6.49 & 22.11 & 29.98 & 0.049 & 0.92 & 0.64 & 0.061 \\
\rowcolor{gray!40} \checkmark&\checkmark&\checkmark& 6.36 & 18.82 & 30.10 & 0.047 & 0.95 & 0.59 & 0.043 \\
    \bottomrule
\end{tabular}
  }
\caption{Ablation study on the reconstruction task verifies the effectiveness of our proposed modules.}
  \label{tab:3}
\end{table*}

\noindent\textbf{Position Control Ability.} Given that the target object is absent in the deletion task, we employ 3D object detection on the remaining three tasks. Table~\ref{tab:2} reveals a slight performance drop in the edited data compared to the unedited oracle data. In terms of translation error, repositioning, insertion, and replacement incur errors of 0.26 meters, 0.24 meters, and 0.32 meters, respectively, with an average translation error of only 0.27 meters. Replacement exhibits a more pronounced performance degradation compared to the other two tasks, primarily attributed to size discrepancies between the replaced and original objects.  All three tasks demonstrate comparable orientation errors, approximately 0.007 radians higher than the oracle. The visualization results in Figure~\ref{fig4} also corroborate the accurate alignment of object positions with GT bounding boxes. These results validate DriveEditor's proficiency in controlling both position and orientation.

\begin{figure}[t]
\centering
\includegraphics[width=\columnwidth]{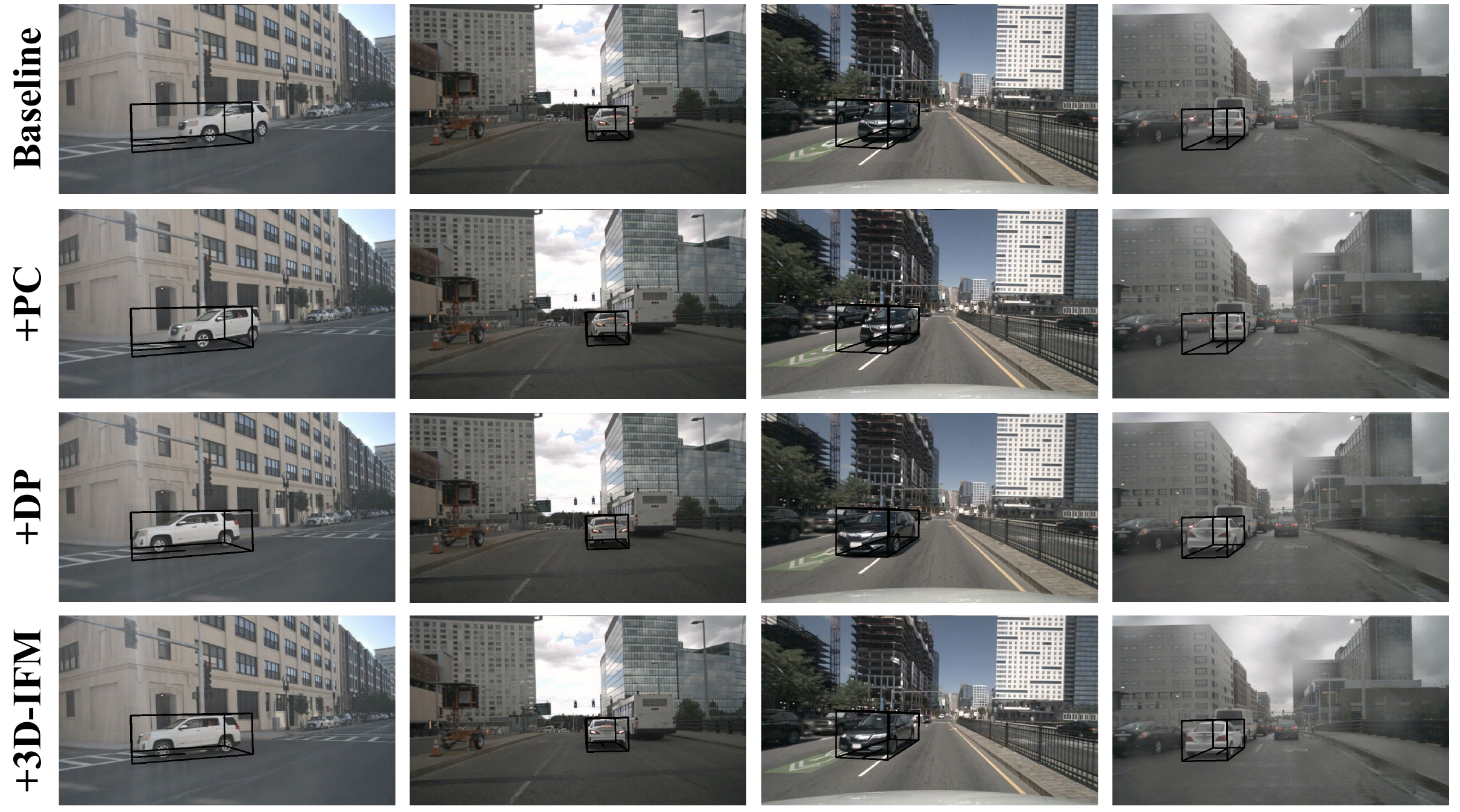}
\caption{Effectiveness of our proposed modules in controlling the position and orientation of objects. GT bounding boxes are outlined in black within the images.}
\label{fig5}
\end{figure}

\begin{figure}[t]
\centering
\includegraphics[width=\columnwidth]{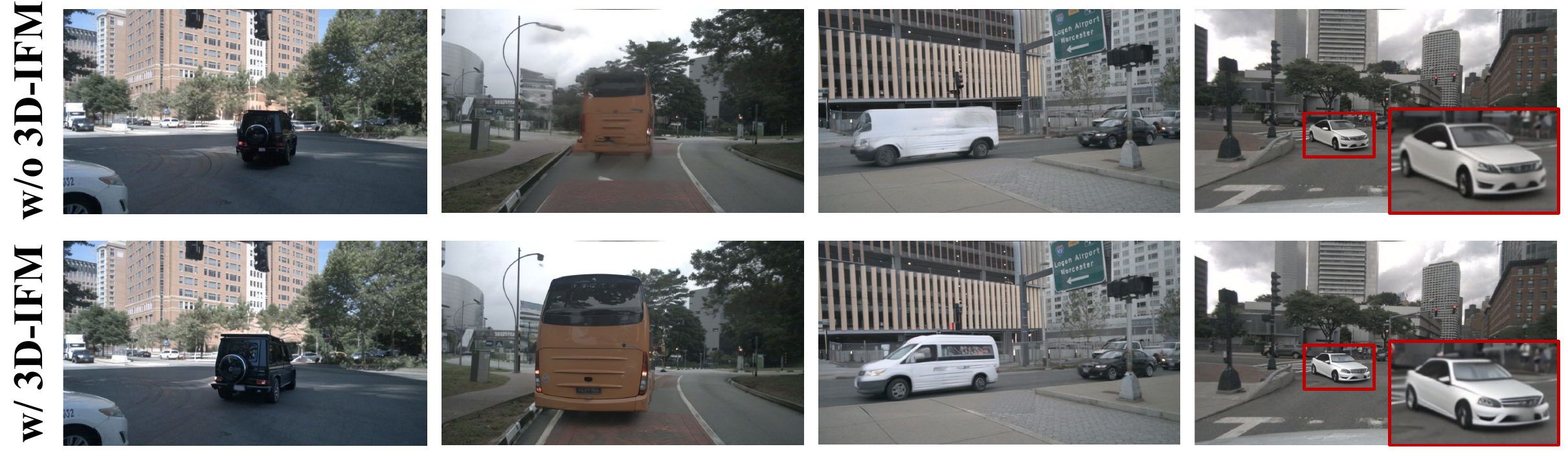}
\caption{Effectiveness of the 3D information fusion module in preserving object appearance and preventing distortion.}
\label{fig6}
\end{figure}

\subsection{Ablation Study}
We conduct a comprehensive ablation study on the reconstruction task, evaluating both editing quality and position control ability, as presented in Table~\ref{tab:3}.

As a \textit{baseline}, we ablate the 3D information fusion and position control modules, forcing the model to rely solely on the single cut-and-paste frame and mask positions to infer object position. This results in a significant translation error of 0.74 meters and an orientation error of 0.197 radians.

To access our Pose Controller (\textit{PC}), we project 3D bounding box edges onto image plane and use it to encode non-depth positional features. While this approach yields improvements in positional metrics, it also introduces subtle degradations in quality compared to the baseline, potentially due to interference from the injected positional information.

We then introduce Depth-aware Projection (\textit{DP}), which yields a significant improvement of 0.087 radians in mAOE and 0.06 meters in mATE.  The remarkable improvement in quality metrics compared to the baseline further underscores the importance of depth information.

Finally, we incorporate the 3D priors through 3D Information Fusion Module (\textit{3D-IFM}), to develop DriveEditor. This integration significantly enhances video quality and consistency, as evidenced by FID, FVD, and PSNR scores of 6.36, 18.82, and 30.10, respectively. Notably, mRecall improves substantially from 0.92 to 0.95, indicating a reduction in unexpected object appearance distortion, as visually confirmed in Figure~\ref{fig6}. Concurrently, position control is improved, with mATE and mAOE decreasing by 0.05 meters and 0.018 radians, respectively, as visualized in Figure~\ref{fig5}.

\begin{table}[t]\small
  \centering
  \setlength{\tabcolsep}{1mm}{
\begin{tabular}{>{\centering\arraybackslash}m{0.6cm} >{\centering\arraybackslash}p{1.2cm} >{\centering\arraybackslash}p{1.2cm}|cccc}
    \toprule
  \multirow{2}[0]{*}{\centering Real} & Gen 50\% \newline Repo. & Gen 50\% \newline Repl.& \multirow{2}[0]{*}{\centering mAP $\uparrow$} & \multirow{2}[0]{*}{\centering mATE $\downarrow$} & \multirow{2}[0]{*}{\centering mAOE $\downarrow$} & \multirow{2}[0]{*}{\centering NDS $\uparrow$} \\
    \midrule[0.5pt]
 \checkmark & - & - & 0.480 & 0.615 & 0.378 & 0.569 \\
 \checkmark & \checkmark & - & 0.482 & 0.600 & 0.340 & 0.577 \\
 \checkmark & - & \checkmark & 0.487 & 0.588 & 0.375 & 0.576 \\
\rowcolor{gray!40} \checkmark&\checkmark&\checkmark& 0.488 & 0.582 & 0.338 & 0.581 \\
    \bottomrule
\end{tabular}
}
\caption{Comparison of augmented datasets generated via repositioning (\textit{Repo.}) and replacement (\textit{Repl.}), each derive from the same 50\% subset of the nuScenes training set. The augmented data significantly benefits downstream tasks.
}
  \label{tab:4}
\end{table}

\subsection{Training Support for 3D Object Detection}
Since not all data contains suitable objects for editing, we select 50\% of the nuScenes training data to generate two augmented datasets through repositioning and replacement, respectively, to enhance StreamPETR model training. As shown in Table~\ref{tab:4}, repositioning expands the object viewpoint distribution, resulting in a 0.038 radian reduction in orientation error compared to non-augmented data. Replacement introduces diverse objects at identical positions, leading to a decrease in translation error. Combining both augmented datasets, despite being derived from the same original data, significantly reduces translation and orientation errors, achieving a NuScenes Detection Score (NDS) of 0.581. This demonstrates that DriveEditor can significantly enrich the diversity of data, thereby facilitating downstream tasks.

\section{Conclusion}
This paper introduces DriveEditor, a diffusion-based unified framework that allows users to easily reposition, insert, replace, and delete objects within driving scenario videos. Additionally, it supports iterative editing by conditioning on the last frame of the previous video, allowing for the editing of long videos. With our proposed position control modules, DriveEditor achieves alignment of edited results with 3D bounding boxes, enabling highly controllable position manipulation. Besides, DriveEditor preserves appearance through different feature levels, enabling high-fidelity object appearance control based solely on a single reference image. Through extensive experiments, DriveEditor has demonstrated exceptional fidelity and controllability in object editing within driving scenarios.

\section*{Acknowledgements}
This work was supported by grants from the National Natural Science Foundation of China (62176100, 62301228, 62376011). The computation is completed in the HPC Platform of Huazhong University of Science and Technology.

\bibliography{aaai25}

\appendix
\newpage
\twocolumn[
\begin{@twocolumnfalse}
\begin{center}
\textbf{\LARGE Appendix}
\vspace{2em}
\end{center}
\end{@twocolumnfalse}
]

\section*{Summary}
This appendix is organized as follows.
\begin{itemize}
    \item Section 1 provides more implementation details.
    \item Section 2 provides more details of the experiments.
    \item Section 3 provides more visualization results.
    \item Section 4 provides the limitations of our work.
\end{itemize}

\section{More Implementation Details}
\subsection{Azimuth Angles in 3D Prior Incorporation}
\begin{figure}[t!]
    \centering
    \begin{subfigure}[b]{\columnwidth}
        \centering
        \includegraphics[width=\columnwidth]{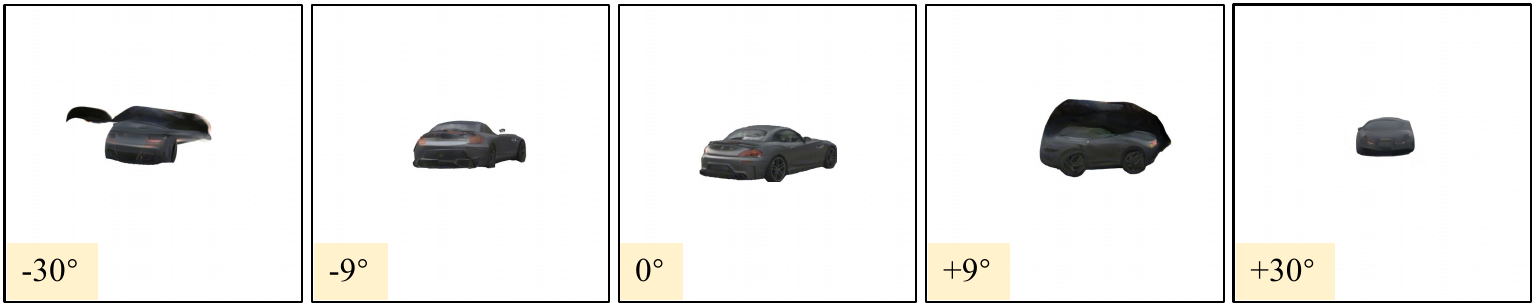}
        \caption{Generate 14 novel-views without full-circle trajectory.}
        \label{sup1:subfig1}
    \end{subfigure}
    \begin{subfigure}[b]{\columnwidth}
        \centering
        \includegraphics[width=\columnwidth]{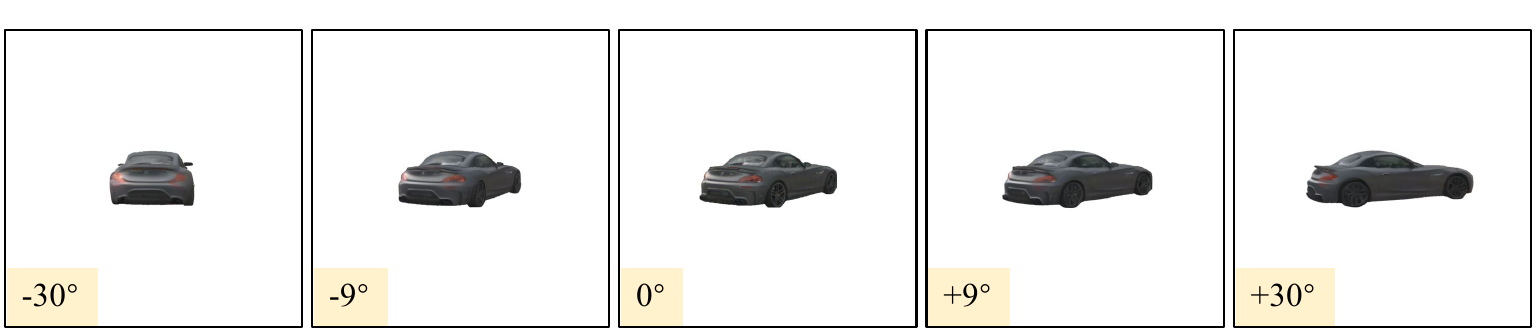}
        \caption{Generate 21 novel-views with our fixed set of azimuth angles.}
        \label{sup1:subfig2}
    \end{subfigure}
    \caption{Novel view images generated by the SV3D model. The corresponding angle of view change is indicated in the bottom left corner of each image.   Our predefined set of azimuth angles produces higher quality novel views compared to those generated from a non-full-circle trajectory.}
    \label{fig:sup1}
\end{figure}

\begin{figure}[t!]
\centering
\includegraphics[width=0.78\columnwidth]{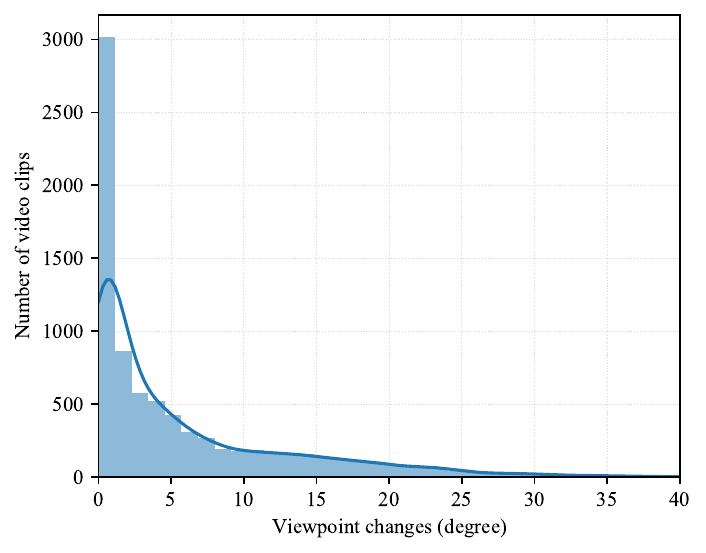}
\caption{Angular distribution map of viewpoint changes of object within our training videos. As shown in the figure, the viewpoint changes of the target object in most videos are only a few degrees.} 
\label{fig:sup2}
\end{figure}

Considering the relatively low speed of object movement in driving scenarios, the viewpoint changes of the target object within videos are relatively small. We have analyzed all video clips in the training set and found that the average viewpoint change is only 5.67 degrees, refer to Figure~\ref{fig:sup2} for details. However, directly inputting the azimuth angles of objects in each frame of the video into the SV3D model to generate corresponding view images is infeasible, as the SV3D model's trajectory control struggles with handling these fine-grained view changes and non-full-circle trajectories. Additionally, given that it was trained on generating 21 novel-view images, the model encounters difficulties when tasked with generating a smaller sequence of images, as shown in Figure~\ref{sup1:subfig1}. Consequently, the corresponding intermediate features also degrade, failing to provide high-quality priors. To address the aforementioned issue without compromising the model's strong prior, we keep the model frozen and adopt a fixed set of azimuth angles $\tilde{\textbf{a}}=$ [0, 3, 6, 9, 12, 16, 23, 30, 45, 90, 135, 225, 270, 315, 330, 337, 344, 348, 351, 354, 357], to control viewpoints. This approach yields high-quality novel view results, as illustrated in Figure~\ref{sup1:subfig2}. For each video frame, we calculate the angular difference between the object's viewpoint and that of the reference image. The viewpoint within the SV3D model that most closely matches this computed angle is subsequently selected as the target frame. Features extracted from these target frames are injected into the corresponding frames of the video. Our fixed viewpoint generation strategy is capable of providing high-quality novel views while achieving an average angular difference of only 1.08 degrees.

\subsection{Position Control}
\begin{figure}[t]
\centering
\includegraphics[width=\columnwidth]{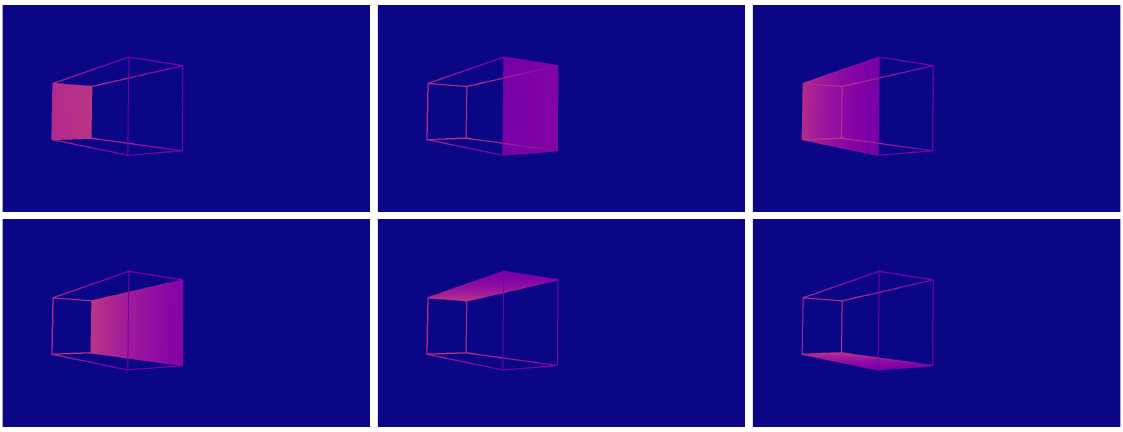}
\caption{Visualization of our depth-aware position images. Areas with a depth of 0 are mapped to blue.}
\label{fig:sup3}
\end{figure}

\begin{figure}[t]
\centering
\includegraphics[width=0.85\columnwidth]{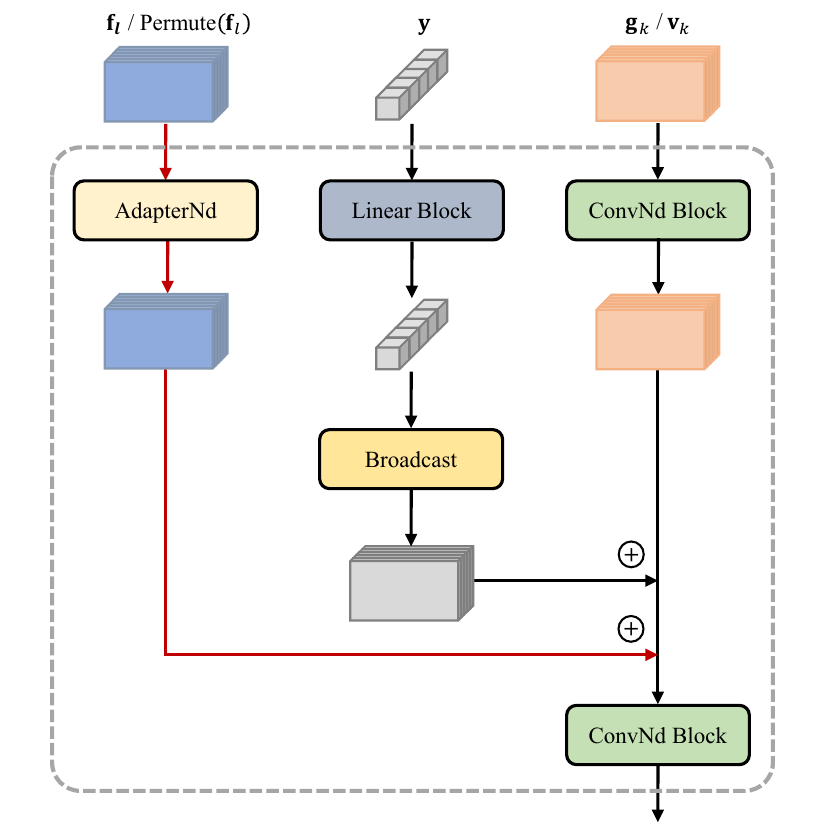}
\caption{The injection of position features into the k-th spatial/temporal ResBlock of the video model. The newly added path is highlighted in red.}
\label{fig:sup4}
\end{figure}

In Depth-aware Projection, we project edges and individual faces of the 3D bounding box onto the image plane, assigning their corresponding depth values $z$ as pixel intensities. Let $\left(u, v\right)$ denote the 2D coordinates in the image plane, $\left(x, y, z\right)$ the 3D coordinates in the camera coordinate system, and $K$ the camera intrinsic matrix. The pixel value at coordinate $\left(u, v\right)$, denoted as $F \left(u, v\right)$, can be represented by the following equation:
\begin{equation}
F\left( \textbf{u}, \textbf{v}\right) = \textbf{z}, \text{where} \begin{bmatrix}\mathbf{u}\\\mathbf{v}\\\mathbf{1}\end{bmatrix} = \mathbf{K} \begin{bmatrix}\mathbf{x}/\mathbf{z}\\\mathbf{y}/\mathbf{z}\\\mathbf{1}\end{bmatrix}
\end{equation}
Through this process, we obtain six projected images, visualized as Figure~\ref{fig:sup3}. These images, arranged in the order of front, back, left, right, top, and bottom, are concatenated along the channel dimension and fed into the Position Controller for encoding. The resulting features are then incorporated into the spatial and temporal ResBlocks of the video U-Net via adapters, as illustrated in Figure~\ref{fig:sup4}.

\subsection{More Model Setup}
DriveEditor leverages pre-trained weights from SVD and SV3D-p for its video and 3D components, respectively. New parameters, except for the zero-init modules, are randomly initialized. The experiments are conducted on 8 NVIDIA A100 GPUs (40GB) using the DeepSpeed ZeRO Stage 2 engine with a batch size of 1 per GPU. Adam optimizer is used with a learning rate of $1\times10^{-5}$ to train the model for a total of 100K iterations, split into two stages of 60K and 40K iterations each. We set a weight of 5 for the mask region to encourage the model to focus on the areas that need modification.

\section{More Experimental Details}
\subsection{Experimental Setup}
To assess the model's ability to generate high-quality images and control object positions, we apply specific editing criteria to video clips from the validation set and subsequently evaluate the editing performance of DriveEditor and its baseline models. Text prompts for the baseline methods are generated using the BLIP model. Experiments were conducted on five object categories: cars, buses, trucks, trailers, and traffic cones. We automatically acquire editing inputs from the object bank to eliminate human bias.

For \textit{repositioning}, we established criteria based on the analysis of object movement patterns in the training data. Stationary vehicles are simulated to move forward or backward gradually for 1.6 meters, while other stationary objects are moved 1 meter to the left. For moving vehicles, we simulate lane changes by moving them 1.1 meters to the left or right and rotating the vehicle's heading by 3.5 degrees. Alternatively, we simulate acceleration or deceleration by gradually adding an additional 1.6 meters of movement to the original position.

For \textit{replacement}, we select objects from the object bank based on similarity in category, distance, and viewpoint, using the chosen object as the reference image.

For \textit{insertion}, we initially detect object-free regions within the video and subsequently select an object from the object bank that shares similar distance and viewpoint characteristics. This object is then positioned within the identified area. We insert larger objects like vehicles into spacious areas while placing smaller objects like traffic cones in smaller regions.

For \textit{deletion}, we directly remove the target object from the video.

\subsection{Training Support for 3D Object Detection}
To enhance the training of StreamPETR, we select 50\% of the nuScenes training dataset that is suitable for object editing and generate an augmented dataset through replacement and repositioning operations. We adopt the same training strategy as employed in Panacea and utilize the official training parameters of StreamPETR. We first conduct a pre-training phase on the augmented dataset, followed by a fine-tuning stage on the official nuScenes training set.

\section{More Visualization Results}
\subsection{Generalization Performance}
\begin{figure}[t]
\centering
\includegraphics[width=1.\columnwidth]{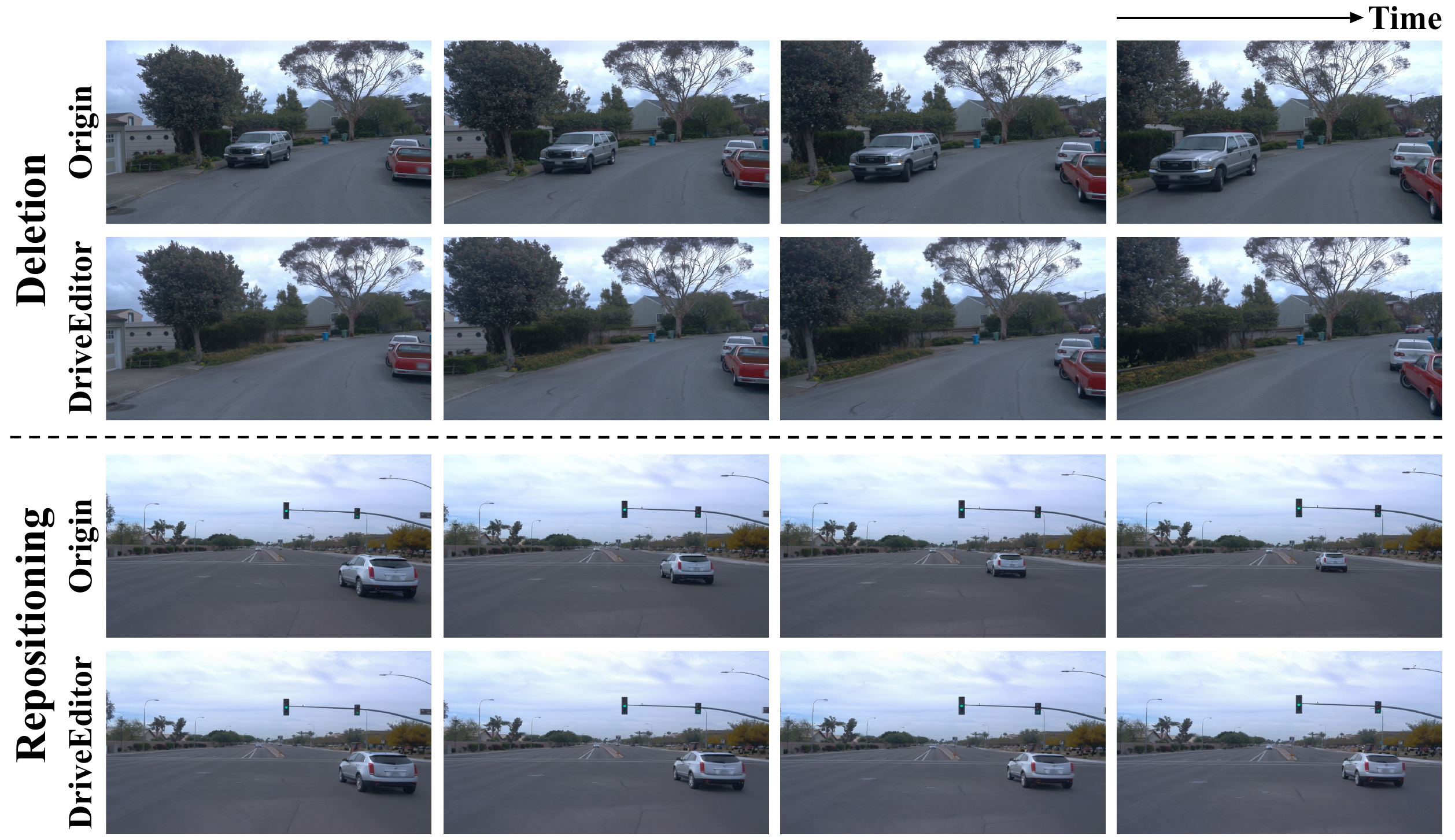}
\caption{Qualitative results of zero-shot experiments on the Waymo dataset demonstrate DriveEditor's strong generalization capabilities despite being trained solely on the nuScenes dataset.}
\label{fig:sup11}
\end{figure}

Despite being trained solely on the nuScenes dataset, DriveEditor exhibits strong generalization capabilities. By focusing on editing only the masked regions of the video while preserving the background and utilizing 3D bounding boxes and reference images to control object position and appearance, DriveEditor demonstrates adaptability to arbitrary driving scenarios. We conducted zero-shot qualitative experiments on the Waymo dataset, which has significant data distribution differences from nuScenes, such as saturation and scene environment. As shown in the figure~\ref{fig:sup11}, DriveEditor demonstrates strong generalization performance on this unseen dataset.

\subsection{Long Videos Editing Results}
\begin{figure*}[!t]
  \centering
  \includegraphics[width=\textwidth]{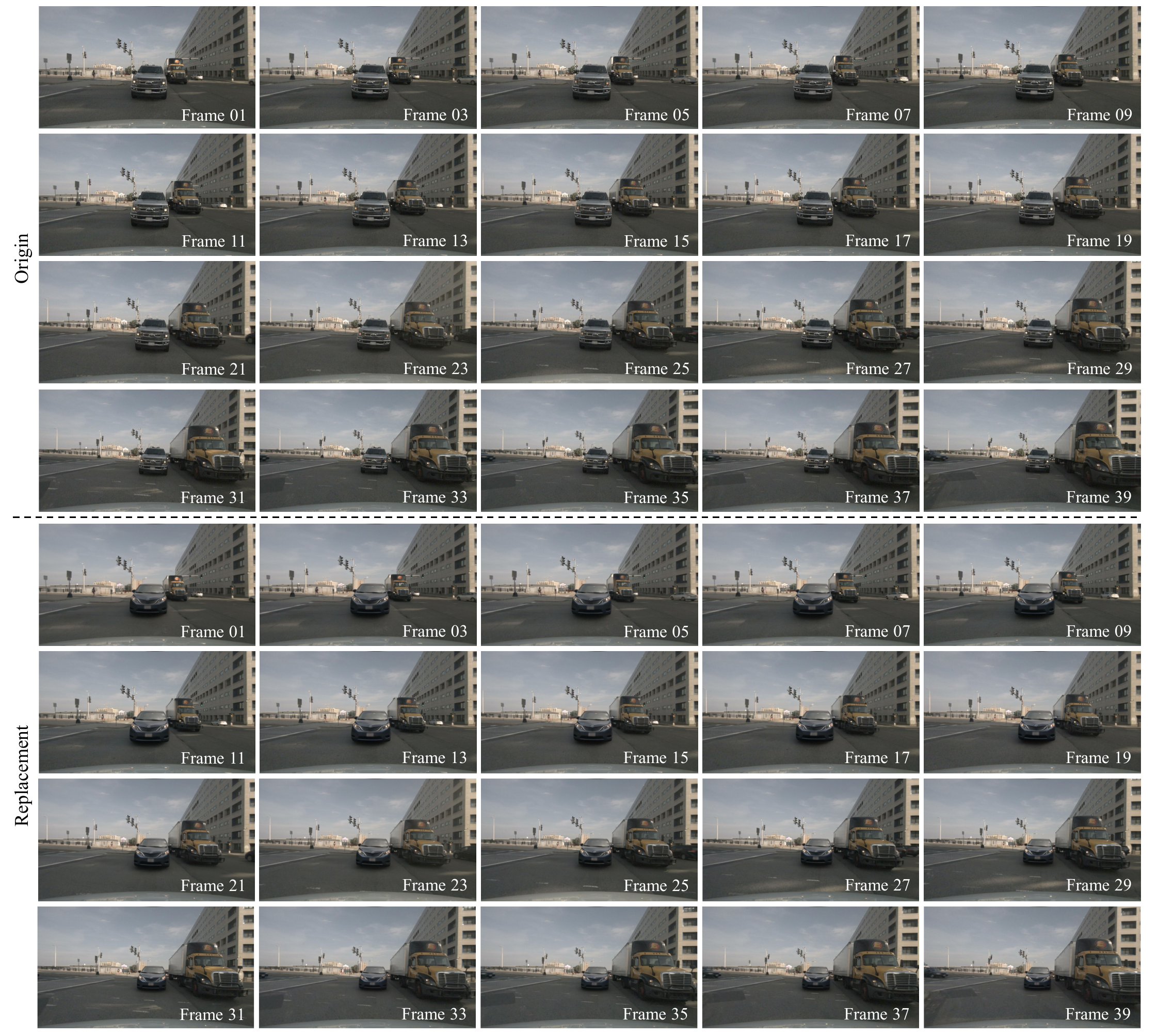}
  \caption{Object replacement results of a 39-frame long video by DriveEditor. }
  \label{fig:sup5}
\end{figure*}

\begin{figure*}[!t]
  \centering
  \includegraphics[width=\textwidth]{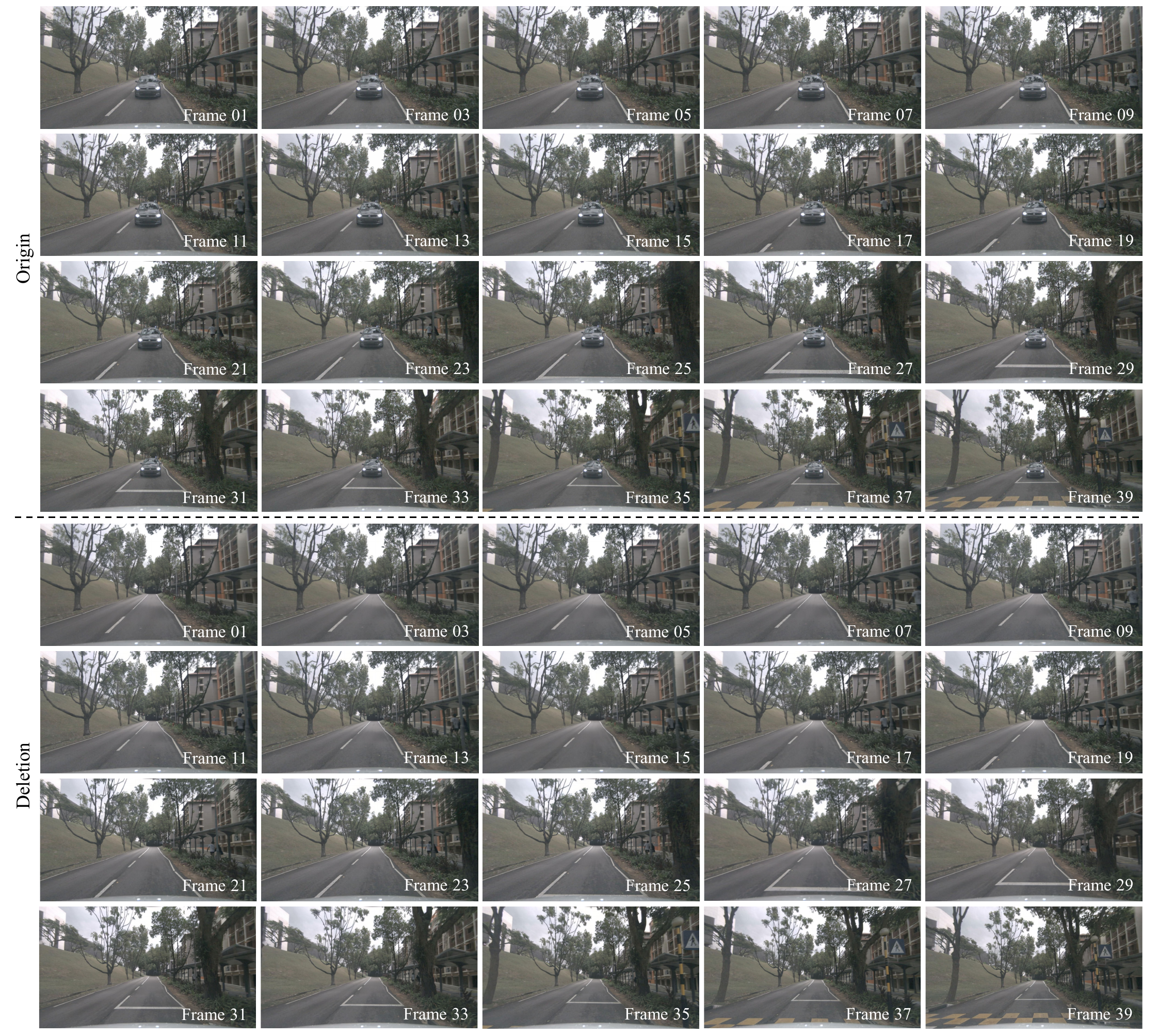}
  \caption{Object deletion results of a 39-frame long video by DriveEditor. }
  \label{fig:sup6}
\end{figure*}

To enhance efficiency, DriveEditor is trained on 10-frame videos but can generate significantly longer sequences. Furthermore, we can iteratively employ the final frame of a preceding video segment as the initial frame for developing the subsequent video segment, instead of using a masked image. As shown in Figure~\ref{fig:sup5} and Figure~\ref{fig:sup6}, we edited a 20-frame video using DriveEditor in a single step, then repeated this process twice and applied iterative editing, resulting in a temporally consistent 39-frame video. This demonstrates DriveEditor's proficiency in editing long videos.

\subsection{More Editing Results}
\begin{figure*}[!h]
  \centering
  \includegraphics[width=0.88\textwidth]{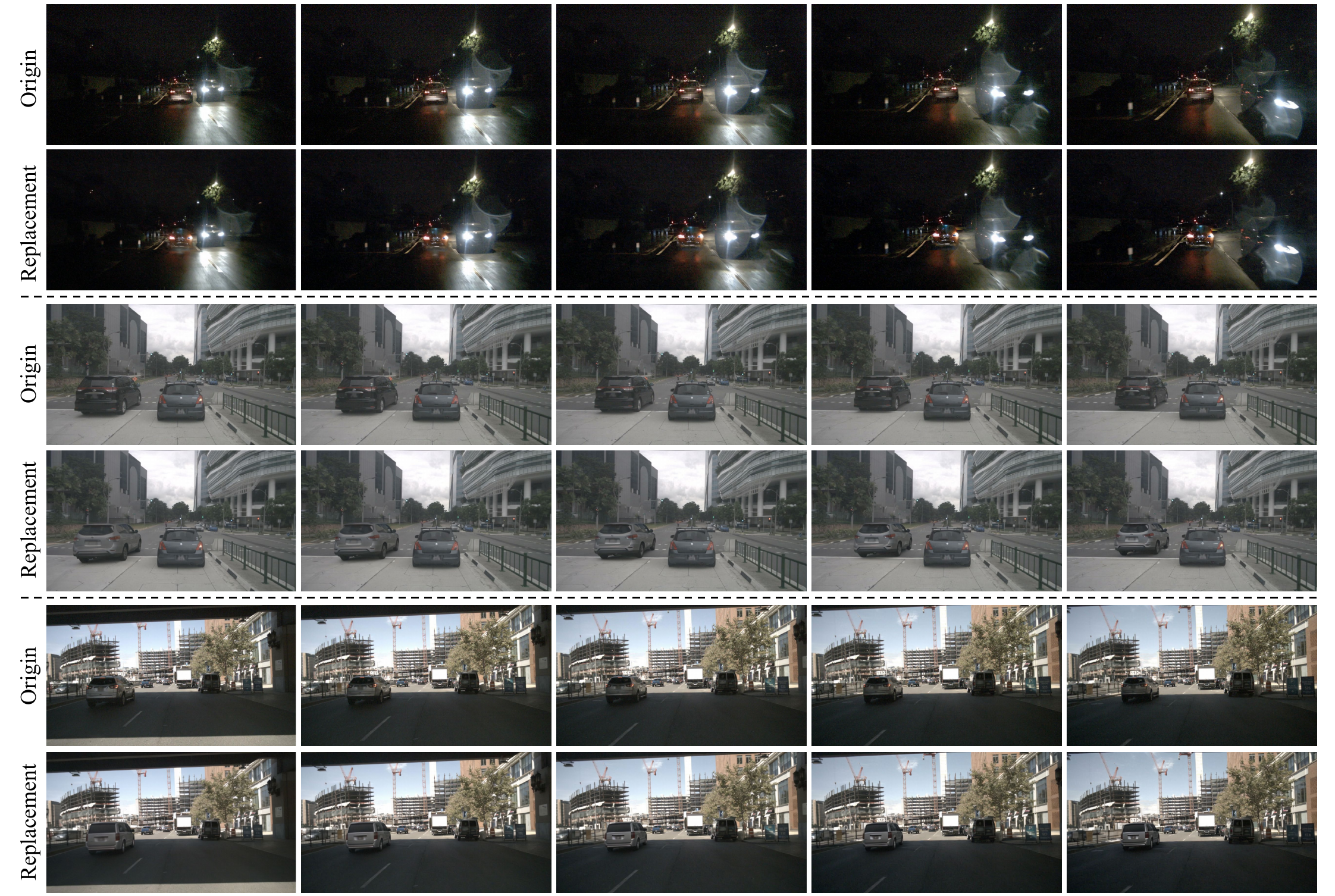}
  \caption{Replacement results by DriveEditor.}
  \label{fig:sup7}
\end{figure*}

\begin{figure*}[!h]
  \centering
  \includegraphics[width=0.88\textwidth]{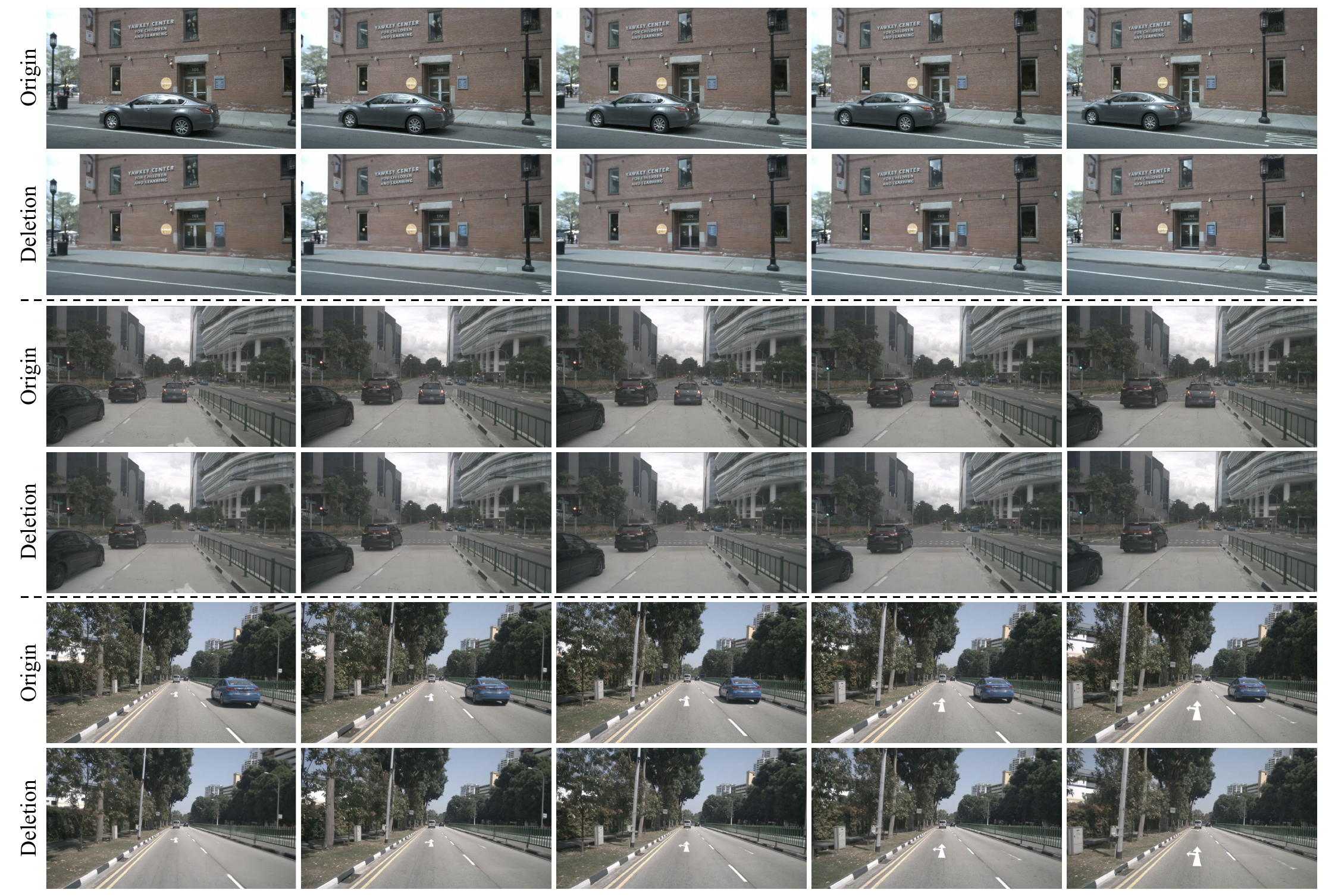}
  \caption{Deletion results by DriveEditor.}
  \label{fig:sup8}
\end{figure*}

\begin{figure*}[!h]
  \centering
  \includegraphics[width=0.88\textwidth]{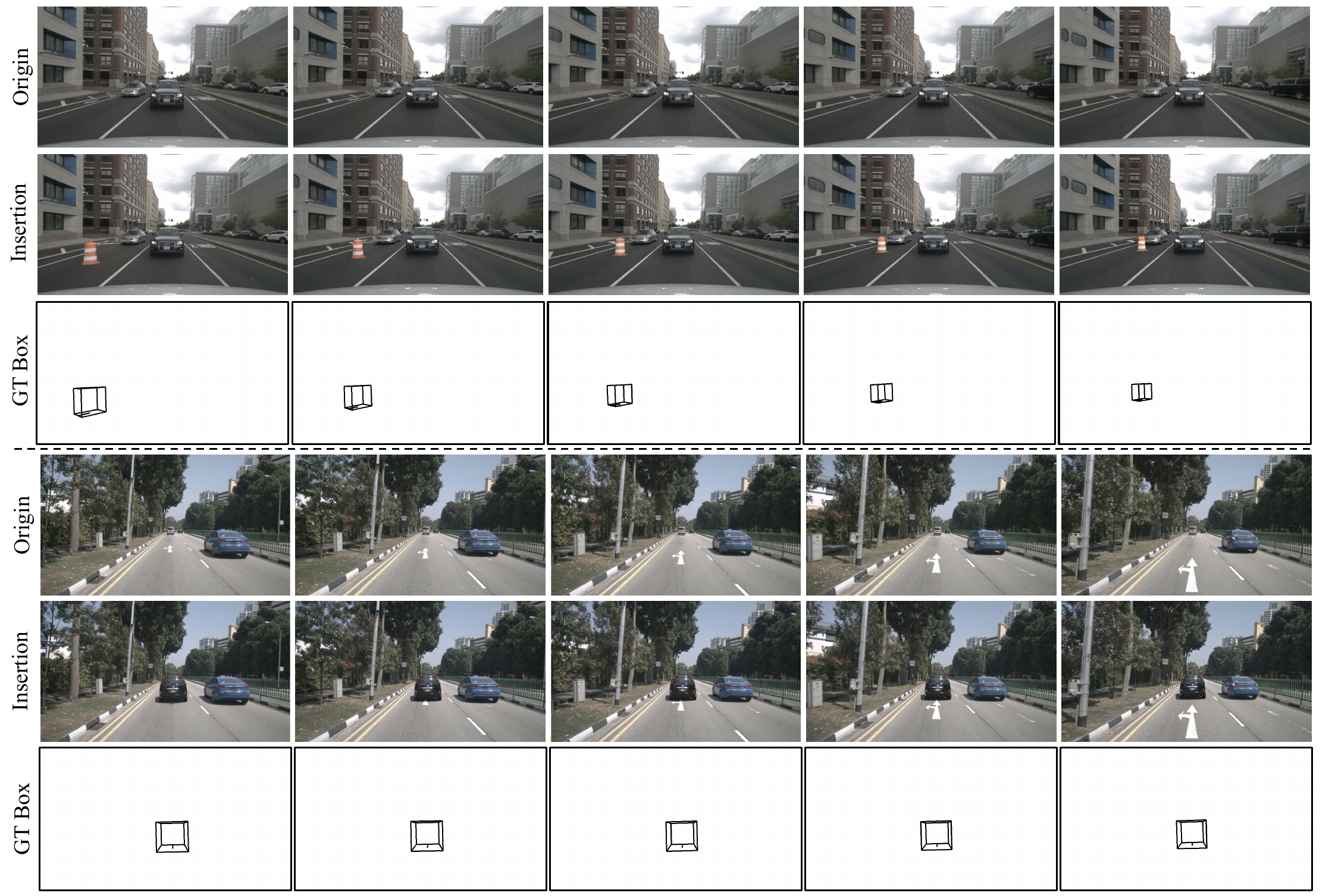}
  \caption{Insertion results by DriveEditor.}
  \label{fig:sup9}
\end{figure*}

\begin{figure*}[!h]
  \centering
  \includegraphics[width=0.88\textwidth]{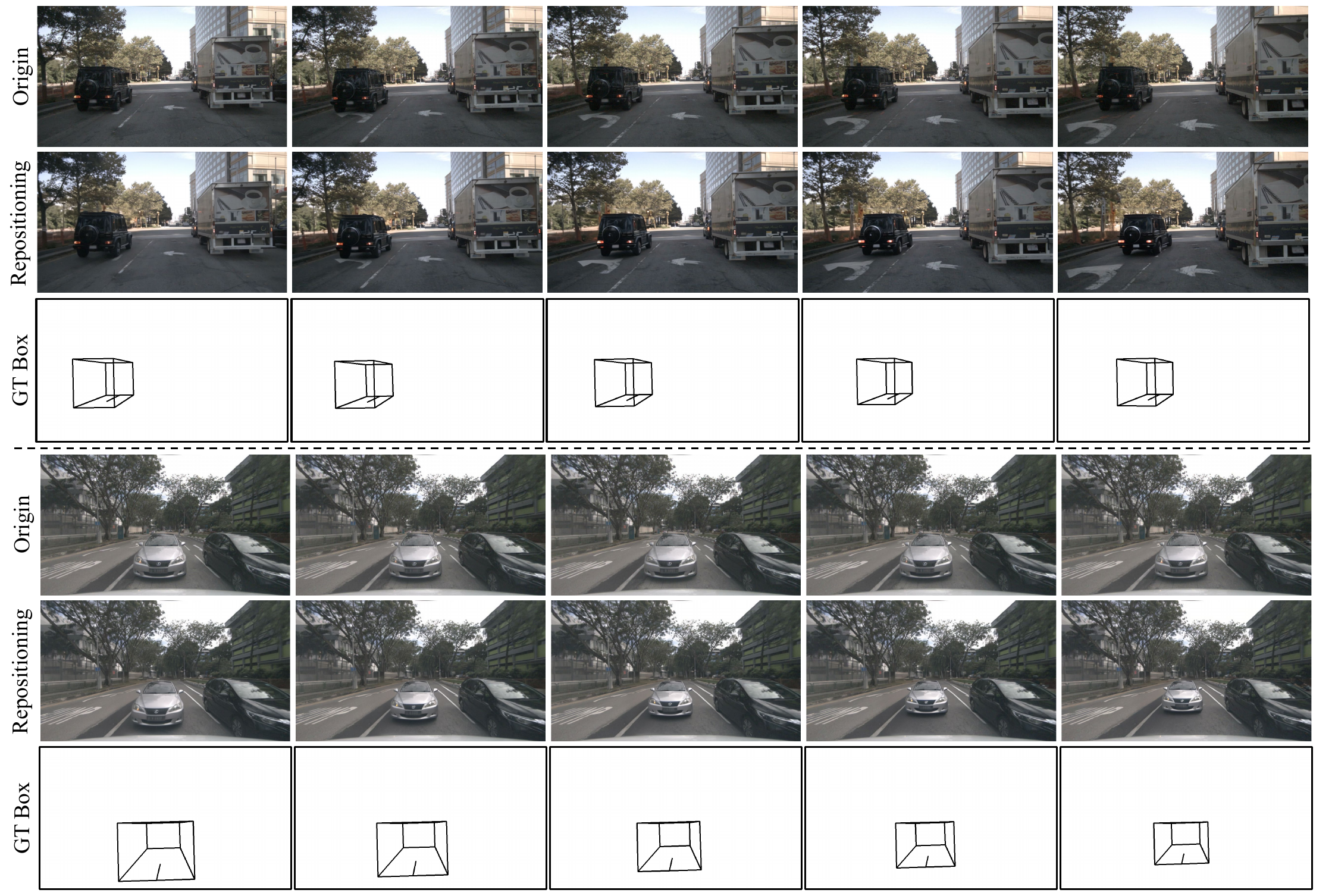}
  \caption{Repositioning results by DriveEditor.}
  \label{fig:sup10}
\end{figure*}

We present additional editing results generated by DriveEditor. Figure~\ref{fig:sup7}-\ref{fig:sup10} respectively show more results of replacement, deletion, insertion, and repositioning results on the validation set. These results further demonstrate that our editing operations apply to a wide range of environments, such as rainy and nighttime conditions. We can achieve high-quality object editing and accurately control object positions. More video results can be found on the project page: https://yvanliang.github.io/DriveEditor.

\section{Limitations}
To achieve unified editing of four diverse object manipulation tasks within a single framework, certain trade-offs have been made. In object repositioning, smaller objects or low-contrast scenes (e.g., rain or nighttime) may lead to blurred and incomplete objects. This issue arises because the object must first be segmented from the source video to serve as a reference image. However, accurate segmentation in such challenging scenarios is difficult. For object insertion and replacement, if the user-specified object significantly differs from the environment (e.g., placing a daytime object into a nighttime scene), the results may exhibit visual distortions. We attribute it to the strong prior imposed by our Appearance Preservation Modules. In future work, we will explore more diverse object appearance manipulation methods, such as implicit feature-based object appearance control.

\end{document}